%% file: nips_2018.tex
\documentclass{article}

\PassOptionsToPackage{numbers, compress, sort}{natbib}
\usepackage[preprint]{nips_2018}
\usepackage[utf8]{inputenc}
\usepackage[T1]{fontenc}   
\usepackage{hyperref}      
\usepackage{url}           
\usepackage{booktabs}      
\usepackage{amsfonts}      
\usepackage{nicefrac}      
\usepackage{microtype}     
\usepackage{verbatim}

\usepackage{algorithm}    
\usepackage{algpseudocode}

\title{Training Verified Learners with Learned Verifiers}

\author{
Krishnamurthy (Dj) Dvijotham\thanks{denotes the primary contributors}\\
DeepMind\\
  \texttt{dvij@cs.washington.edu} \And
Sven Gowal $^\ast$ \\
DeepMind\\
  \texttt{sgowal@google.com} \And
Robert Stanforth $^\ast$\\
DeepMind\\
  \texttt{stanforth@google.com} \And
Relja Arandjelovi\'c \\
  DeepMind\\
  \texttt{relja@google.com} \And
Brendan O'Donoghue \\
  DeepMind\\
  \texttt{bodonoghue@google.com} \And
  Jonathan Uesato \\
  DeepMind\\
  \texttt{juesato@google.com} \And
  Pushmeet Kohli \\
  DeepMind\\
  \texttt{pushmeet@google.com}
}

\usepackage{subfig}
\usepackage{color}
\usepackage{caption}
\usepackage{graphicx}
\usepackage{amsmath}
\usepackage{amsthm}
\usepackage{mathtools}
\usepackage{amssymb}
\usepackage{color}
\usepackage{url}

\renewcommand{\paragraph}[1]{\smallskip\noindent{\bf{#1}}}

\usepackage{tikz}
\usetikzlibrary{shapes,arrows,shadows,fit, backgrounds, decorations.pathmorphing}
\tikzstyle{op} = [draw, fill=red!20, text centered,
                  minimum height=2em, circle]
\tikzstyle{mlp} = [draw, text width=7em, fill=blue!20, text centered,
                  minimum height=2em, rounded corners]
\tikzstyle{loss} = [draw, text width=7em, fill=red!20, text centered,
                  minimum height=2em, rounded corners]
\tikzstyle{var} = [draw, text width=7em, fill=white!20, text centered,
                  minimum height=2em, rounded corners]
\tikzstyle{inp} = [text centered, minimum height=2em]
\tikzstyle{system} = [draw, dotted, minimum height=2em]
\tikzstyle{null} = [inner sep=0, outer sep=0]
\tikzset{>=latex}
\tikzset{every picture/.style={line width=1pt}}

\newcommand{\br}[1]{\left({#1}\right)}

\newcommand{\norm}[1]{\left\| {#1} \right\|}
\DeclareMathOperator*{\argmax}{argmax}
\DeclareMathOperator*{\argmin}{argmin}

\newcommand{\xin}{x_0}
\newcommand{\xnom}{x^\text{nom}}

\newcommand{\lam}[1]{{\lambda}^{{#1}}}

\newcommand{\Sin}{\mathcal{S}_\text{in}}

\newcommand{\dualfun}{\zeta}

\newcommand{\wall}{\mathbf{w}}
\newcommand{\wparam}{\theta}
\newcommand{\D}{\mathcal{D}}
\newcommand{\eg}{{\it e.g.}} 
\newcommand{\ie}{{\it i.e.}} 
\newcommand{\etc}{{\it etc.}}
\newcommand{\reals}{{\mathbb{R}}}

\newcommand{\allx}{\mathbf{x}}
\newcommand{\lambdall}{{\boldsymbol \lambda}}
\DeclareMathOperator*{\maxi}{\text{maximize}}
\DeclareMathOperator*{\mini}{\text{minimize}}

\newtheorem{definition}{Definition}

\begin{document}
\maketitle

\begin{abstract}
This paper proposes a new algorithmic framework, \emph{predictor-verifier training}, to train neural networks that are \emph{verifiable}, \ie, networks that \emph{provably} satisfy some desired input-output properties. The key idea is to simultaneously train two networks: a \emph{predictor} network that performs the task at hand, \eg, predicting labels given inputs, and a \emph{verifier} network that computes a bound on how well the predictor satisfies the properties being verified. Both networks can be trained simultaneously to optimize a weighted combination of the standard data-fitting loss and a term that bounds the maximum violation of the property. Experiments show that not only is the predictor-verifier architecture able to train networks to achieve state of the art \emph{verified} robustness to adversarial examples with much shorter training times (outperforming previous algorithms on small datasets like MNIST and SVHN), but it can also be scaled to produce the first known (to the best of our knowledge) verifiably robust networks for CIFAR-10. 
\end{abstract}

\input{intro.tex}

\input{verification.tex}

\input{pc.tex}

\input{experiments.tex}

\input{conclusions.tex}

\bibliographystyle{abbrvnat}
\bibliography{references}
\input{appendix.tex}

\end{document}

%% file: intro.tex
\section{Introduction} 
Neural networks are increasingly being deployed in a wide variety of applications with great success \citep{krizhevsky2012imagenet,Goodfellow-et-al-2016}. However, recently researchers have raised concerns about the robustness of these models. It has been shown that the addition of small but carefully chosen deviations to the input, so-called adversarial perturbations, can cause the neural network to make incorrect predictions with very high confidence \citep{szegedy2014intriguing, kurakin2017adversarial,carlini2017adversarial, goodfellow2015explaining, carlini2017towards}. 

Starting with the work of \citet{szegedy2014intriguing}, there have been several papers both on understanding adversarial attacks on neural networks and developing defense strategies against these attacks~\citep{warde201611, yuan2017adversarial, akhtar2018threat, xie2018mitigating, liao2018defense, papernot2016distillation}. Recent work has shown that many of the defense strategies proposed in the literature are really obfuscation strategies, that simply make it harder for gradient-based adversaries to attack the model by obfuscating gradients in the model and that these defenses are easily broken by stronger adversaries \citep{uesato2018adversarial, athalye2018obfuscated}. Robust optimization techniques, like those developed by \citet{madry2018towards, MadryRotation}, overcome this problem since they attempt to find the worst case adversarial examples at each step during training which they then augment the training batch with. However, they still do not guarantee that a stronger adversary (for example, one that does brute-force enumeration to compute adversarial perturbations) cannot find inputs where that cause the model to predict incorrectly. Requiring robustness to \emph{any adversarial attack} is an example of a \emph{specification, \ie, a relationship between the inputs and outputs of the network that must hold for all inputs within a given set.} \emph{Verification} refers to the problem of checking a specification, for example, \emph{finding a proof that the model is not susceptible to attack by any adversary}. Algorithms based on formal verification \citep{katz2017reluplex, ElhersPlanet, gehrai} and mixed integer programming \citep{fischetti2018deep, tjeng2017verifying} have been proposed to solve this problem for networks with piecewise linear activation functions (like ReLUs and maxpooling). However, these algorithms can take exponential time (since they perform exhaustive enumeration in the worst case) and are not efficient for large scale machine learning models (the networks considered in these papers are tiny, typically with a few hundred hidden units). 

Even if an efficient verification algorithm is available, it is of limited use if the neural network does not satisfy the specification of interest, since it does not provide any information on how the network should be modified so as to satisfy the specification. Hence, there is a need to go beyond data-driven training of neural networks and towards \emph{verified training}, where neural networks are trained both to fit the data and satisfy a specification \emph{provably, \ie, with guarantees that the specification is not violated even for adversarially chosen inputs}. Recent work on verified training employs ideas from convex optimization and duality to construct bounds on an optimization formulation of verification~\citep{kolter2017provable,CertifiedDefenses}. However, these approaches are limited to either a particular class of activation functions (piecewise linear models) or particular architectures (single hidden layer, as in \cite{CertifiedDefenses}) and furthermore, are computationally intensive since they require solving an optimization problem inside each iteration of the training loop. The limitations on activation functions and architectures were overcome in a recent work by \citet{DjDual}, where an algorithm is derived to compute bounds on the worst case violations of a given network with respect to some specification. However, this work focused on verification of an already trained model and did not consider verified training. Moreover, the method is inappropriate to be used during training due to the high computational expense of solving the optimization problems involved in verifying the neural network. 

Motivated by the need for a scalable and general purpose approach for verified training, we propose a new algorithmic framework, \emph{predictor-verifier training (PVT)}. The predictor is a neural network that performs the task at hand (\eg, classification), while the verifier is a separate neural network that computes a bound on the worst case violation of the PVT we desire the predictor to satisfy. The two networks are jointly trained on a loss function that is a weighted combination of the task loss (that depends only on the predictor) and the dual loss, an upper bound on the worst case violation of the specification (that depends on the predictor weights and the output of the verifier). More concretely, the approach exploits the property of duality based verification algorithms in \citep{DjDual} that shows that \emph{any choice of the dual variables provides a valid upper bound on the worst case violation of the specification}. The approach in \citep{DjDual} chooses the dual variables to optimize this bound. However, in the context of verified training, this introduces significant overhead. Instead, PVT exploits the idea that the solution of this optimization problem can be \emph{learned}, \ie, the mapping from a nominal training example to the optimal dual variables can be learned by the verifier network. Thus, PVT amortizes the cost of verification over all the training examples and alleviates the burden of performing an expensive computation such as solving an optimization problem for each training example, as is required in \citep{kolter2017provable}. Further, since we build upon \citep{DjDual}, we are not limited by the kind of architecture or activation function overcoming the restrictions of the approach in \citep{CertifiedDefenses} (which is limited to single hidden layer networks). Thus, we can outperform \citep{kolter2017provable, CertifiedDefenses} in terms of clean accuracy (in the absence of perturbations), verified adversarial accuracy (a lower bound on the test set accuracy under \emph{any} adversary using bounded magnitude input perturbations) and training time. To summarize, the key contributions of our paper are:\\
(1) A novel training method, predictor-verifier training, that combines training and verification to train accurate models that are consistent with given specifications, which 
    \begin{itemize}    
    \item is general-purpose, \ie, applies to arbitrary feedforward architectures with different kinds of layers (convolutional, batch-norm, ReLU, leaky ReLU, sigmoid, tanh, ELU, \etc),
    \item enables amortization of the cost of verification across multiple examples resulting in computational efficiency. 
    \end{itemize}
(2) Experimental validation of the efficacy of our approach, showing that it
    \begin{itemize}
    \item outperforms previous algorithms for verified training -- achieves state-of-the-art \emph{verified accuracy} (under $\ell_\infty$ adversarial perturbations) for MNIST and SVHN,
    \item can be scaled to verify models for larger datasets like CIFAR-10 and obtain \emph{the first nontrivial verified accuracy bounds, to the best of our knowledge.}
    \end{itemize}

%% file: verification.tex
\section{Verification using duality}
We provide a self-contained recap of the verification approach developed in \citep{DjDual}, since we will build on it to develop PVT (section \ref{sec:pc}). 
We focus on neural networks trained for classification tasks, though other tasks can be handled similarly. In classification, the neural network is fed an input $\xin$ (\eg, an image) and outputs a vector of un-normalized log-probabilities (hereafter \emph{logits}) corresponding to its beliefs about which class $\xin$ belongs to. Generally speaking, the network's prediction is taken to be the argmax of the output vector of logits.  During training, the network is fed pairs of inputs and correct output labels, and trained to minimize a misclassification loss, such as the cross-entropy. 

\subsection{The verification problem}

We are concerned with \emph{verifying} that networks satisfy some \emph{specification}, and generating a proof, or \emph{certificate}, that this holds. We consider a general class of specifications that require that for all inputs in some nominal set, the network output is contained in a half-space:
\begin{align}
  c^T \phi(\xin) + d \leq 0, \quad \forall \xin \in \Sin,
 \label{eq:spec}
\end{align}
where $c$ and $d$ define the specification to be verified and $\phi$ is the mapping performed by the neural network. 

As shown in \citep{DjDual}, many useful verification problems fit this definition (monotonicity with respect to certain inputs, ensuring that an unsafe label is never predicted when the inputs lie in a given set \etc). However, in this paper, we will focus on \emph{robustness to adversarial perturbations}. In the adversarial setting, an adversary perturbs a nominal input, $\xnom$, in some set, $\Sin(\xnom)$, and feeds this perturbed input into the network. For example, if the adversary can perturb the input in $\ell_\infty$ norm up to some $\epsilon \geq 0$ then $\Sin(\xnom) = \{x \mid \|x - \xnom\|_\infty \leq \epsilon \}$. In the rest of the paper, we will assume that $\Sin(\xnom)$ is of this form, although the framework can be extended to other constraints on the input.

Adversarial robustness is achieved if no adversary can succeed in changing the classification outcome away from the true label $y^\text{nom}$, \ie, there is no element $x^0$ in $\Sin(\xnom)$ such that $\argmax(\phi(x^0)) \neq y^\text{nom}$. In this case we want to verify that for each label $i$:
\begin{align}
  (e_i - e_{y^\text{nom}})^T \phi(\xin) \leq 0, \quad \forall \xin \in \Sin\br{x^\text{nom}},
 \label{eq:spec2}
\end{align}
where $e_i$ is the $i$th standard basis vector.

We consider a layered architecture with layers numbered $0, \ldots, K-1$, where the input of layer $k$ (and consequently the output of layer $k-1$) is $x_k \in \reals^{n_k}$, with $x_0$ corresponding to the inputs, and $x_K$ denoting the final output, \ie, the logits.
Each layer $k$ has a transfer function, $h_k: \reals^{n_k} \rightarrow \reals^{n_{k+1}}$, that maps its inputs to outputs. We denote by $\wall$ all the trainable parameters in the network and by $\allx$ the collection of activations at all layers of the network $\allx=(x_0, x_1, \ldots, x_K)$.
Searching for a counter-example to \eqref{eq:spec} can then be written as an optimization problem:
\begin{align}
\begin{split}
\maxi_{\allx} \quad & c^T x_K + d \\
\text{subject to} \quad & x_0 \in \Sin\br{x^\text{nom}} \\
& x_{k+1} = h_k(x_k), \quad k = 0, \ldots, K-1.\\
\end{split}
\label{eq:verifyOpt}
\end{align}
If the optimal value of the above optimization problem is smaller than $0$, the specification \eqref{eq:spec} is true.

\subsection{Duality based verification}
\label{sec:dual_verification}

In this section we present an efficient verification method developed in \citep{DjDual} based on a dual relaxation of problem (\ref{eq:verifyOpt}).
We start by outlining the relevant assumptions.

\paragraph{Supported transfer functions.}
The verification method restricts the transfer functions to be \emph{simple}, where we define \emph{simple} as follows:
\begin{definition}\label{def:simple}
A function $h : \reals^n \rightarrow \reals^m$ is \emph{simple} if one can efficiently solve
\begin{equation}
\label{e-simple}
\begin{array}{ll}
\displaystyle\maxi & \mu^T x - \lambda^T h(x) \\
\text{subject to} & l \leq x \leq u,
\end{array}
\end{equation}
over variable $x \in \reals^n$, for fixed $\lambda \in \reals^m$, and $\mu, l, u \in \reals^n$.
\end{definition}
We will also assume that the transfer functions are amenable to backpropagation with respect to any parameters they may have. The above definition uses the term `efficiently solve', which is intentionally ambiguous but could be taken to mean that the problem has a closed-form solution, or can be solved using convex optimization \etc; we just assume access to a cheap oracle for the problem.
In that case this definition covers a wide range of commonly used functions, including most of those used in modern network architectures. For example, it includes
any affine function, \eg, $h(x) = Wx + b$, since in this case the problem \eqref{e-simple} has a closed form solution. Furthermore, any component-wise non-linearity falls into this category, since in that case problem \eqref{e-simple} is decomposable into several independent scalar problems, which in the worst case we could solve by discretizing the interval.
Other functions like max-pool also fall into this category, since in that case the solution to problem \eqref{e-simple} can be computed by solving small problems each of which has a closed-form solution, and taking the maximum \citep{DjDual}.
Consequently, this formulation covers most state-of-the-art (non-recurrent) deep learning architectures, since affine transfer functions include convolutions, average pooling, batch normalization (once the parameters are fixed) \etc, and the non-linearities can include ReLU, sigmoid, tanh and others. Similarly, skip connections, as popularized in the ResNet architecture \citep{he2016deep}, can be formulated in this way by judicious choice of weight matrix and non-linearity.
\paragraph{Bound propagation.}
In this paper we work with the set $\Sin\br{\xnom}=\{x: \norm{x-\xnom}_\infty \leq \epsilon\}$ for some $\epsilon > 0$. Then the upper and lower bounds on $x_0$ are simply $\xnom - \epsilon \leq x_0 \leq \xnom + \epsilon$. The following simple bound propagation approach can then be used to find bounds on subsequent layers: At layer $k$, for each output index $i=1, \ldots, n_{k+1}$, the following pair of problems is solved:
\begin{equation}
\label{e-bounds}
\begin{array}{ll}
\mbox{max / min} & e_i^T h_k(x) \\
\mbox{subject to} & l_{k} \leq x \leq u_{k}\\
\end{array}
\end{equation}
over variable $x \in \reals^{n_k}$, where $e_i$ is the $i$th standard basis vector, and $u_{k}, l_{k}$ are the upper and lower bounds on the outputs from the previous layer. The resulting maximum and minimum values are the upper and lower bounds, respectively, for the inputs to the next layer. Problem \eqref{e-bounds} can be solved efficiently due to the assumption that the transfer function defining each layer is \emph{simple}. For example, if $h$ is a \emph{non-decreasing} component-wise non-linearity (\eg, ReLU), then the bounds at the output are simply $l^\prime = h(l), u^\prime = h(u)$. When $h$ is the affine function $h(x) = Wx + b$, then the output bounds are given by:
$l^\prime = W^+ l + W^- u + b, u^\prime = W^+ u + W^- l + b$
where $W^+=\max\br{W, 0}$ and $W^-=\min\br{W, 0}$. These bounds can be computed at the same time as a forward pass through the neural network at a constant overhead and hence are efficient to compute.

\paragraph{Verification.}
The verification problem \eqref{eq:verifyOpt} is NP-hard so computing the exact optimal value in general is difficult \cite{katz2017reluplex}. However, in order to verify the specification \eqref{eq:spec}, it suffices to prove that the optimal value is bounded above by $0$. Thus, \citet{DjDual} \emph{relax} the optimization problem \eqref{eq:verifyOpt} by introducing Lagrangian multipliers corresponding to the equality constraints, to obtain the problem
\begin{align}
\begin{split}
\dualfun\br{\lambdall; \xnom, \wall} = \maxi_{\allx} \quad & c^Tx_K + d + \sum_{k=0}^{K-1} \lambda_k^T(x_{k+1} - h_k(x_k))\\
\mbox{subject to} \quad & x_0 \in \Sin\br{x^\text{nom}} \\
&
l_k \leq x_k \leq u_k, \quad k = 1, \ldots, K
\end{split}
\label{eq:verifyOptdual}
\end{align}
assuming any fixed values for $\lambda_k$. The optimal value as a function of $\lambdall=\{\lambda_k,\ k=0, \ldots, K-1\}$ is known as the \emph{dual function} and denoted $\dualfun\br{\lambdall; \xnom, \wall}$, acknowledging its dependence on $\xnom, \wall$. Notice how the known bounds on $x_k$ are introduced into this formulation. The introduction of valid constraints does not make any difference to the solution, but the bounds help make the Lagrangian relaxation tighter and compute better upper bounds on the optimal value of \eqref{eq:verifyOpt}. This optimization problem over $\allx$ can be solved in closed form based on the assumption that the transfer functions are simple -- we leave the details for Appendix \ref{sec:AppCF} noting that they are similar to \cite{DjDual}. For any choice of $\lambdall$, the objective value provides an upper bound on the optimal value of problem \eqref{eq:verifyOpt}. A sufficient condition to verify \eqref{eq:spec} is finding a $\lambdall$ such that dual function is non-positive, in other words:
\begin{align}
 \exists \lambdall \enskip \text{such that} \dualfun\br{\lambdall; x^\text{nom}, \wall} < 0 \implies c^T \phi(\xin) + d \leq 0, \quad \forall \xin \in \Sin\br{x^\text{nom}},
 \label{eq:dualfun}
\end{align}
and the dual variable provides a \emph{certificate} that the property holds. On the other hand, if we find an $x_0 \in \Sin(\xnom)$ such that the optimal value of \eqref{eq:verifyOpt} is positive, then we have found a counter example which is as a certificate that the property does not hold. Note that the dual function $\dualfun(\cdot; \xnom, \wall)$ is convex, and so minimizing it can be done efficiently \cite{boyd2004convex}. However, we only need to find a $\lambdall$ for which the objective value is smaller than zero, so it is not required to minimize the function exactly. This motivates the approach of \citet{DjDual}, where the authors apply a subgradient method which for all steps maintains a valid upper bound on the optimal value of \eqref{eq:verifyOpt} that gets progressively tighter.

The most important properties that we use in the following sections are that \emph{any} choice of $\lambdall$ provides a valid upper bound on \eqref{eq:verifyOptdual}, and if that upper bound is smaller than zero then that $\lambdall$ is a valid certificate to verify that the property holds. We shall exploit this to train a neural network to output the dual variables, for a particular input $x_0$. If the dual variables were required to satisfy constraints to be valid (this is the case with the formulation in \citep{kolter2017provable}), or if they needed to exactly optimize an objective to be a certificate (as in the case of exhaustive search methods), then a neural network, which is approximate by its very nature, would not be able to produce them. Further, the dual objective $\zeta$ is (sub)-differentiable with respect to $\lambdall$ and $\wall$, which allows backpropagation to be used to train neural networks to produce near-optimal dual variables.

%% file: pc.tex
\section{Predictor-Verifier Training}\label{sec:pc}
\begin{algorithm}[t]
\caption{Predictor-Verifier Training}
\begin{algorithmic}
\State {\bf input:} Dataset $\D$; loss weighting $\kappa$; Stepsizes $\{\beta_t\}_{t=0}^\infty$
\State {\bf initialize:} Neural network parameters $\theta^0$ (of the verifier), $\wall^0$ (of the predictor) randomly
\For{time-step $t=0,1,\ldots,$}
\State Sample input minibatch $\{(x^i, y^i)\}_{i=1}^m \subseteq \D$
\State Compute loss function estimate $\hat{L}$ using \eqref{eq:loss} on minibatch.
\State $\theta^{t+1} \gets \theta^t - \beta_t \frac{\partial \hat{L}^t}{\partial \theta}$ ~~~~~(Improve verifier to reduce verified dual bound)
\State $\wall^{t+1} \gets \wall^t - \beta_t \frac{\partial \hat{L}^t}{\partial \wall}$ ~~(Improve the predictor and make it more verifiable)
\EndFor
\end{algorithmic}
\label{alg:PVT}
\end{algorithm}
In this section, we discuss how the verification approaches from the previous section can be folded into neural network training approaches in order to learn networks that are guaranteed to satisfy the specification. The key idea is to exploit the fact that the dual optimization problem shares a lot of structure across training examples, so that the solution of the optimization problem
$\lambdall^\star\br{x, \wall}=\argmin_{\lambdall} \dualfun\br{\lambdall;x, \wall}$
can be ``learned'', \ie, a neural network can be trained to approximate the optimal solution $\lambdall^\star$ given $x, \wall$. This alleviates the burden of solving the above optimization problem within each iteration in the training loop. This is done by using a verifier network $V$ to predict the dual variables given the input example. As the network trains, the verifier learns to produce dual variables that approximately minimize the upper bound (\ie, provide tighter bounds) and the predictor adjusts its weights so that the violation of the  specification is minimized.
The entire process is amenable to backpropagation and training with standard stochastic gradient algorithms (or variants thereof), making the training of verifiable models very similar to regular training. Concretely, we train two networks simultaneously:\\
1) A predictor $P$, that takes as input the data to be classified $x_0$ and produces the logits as output. The predictor is parameterized by $\wall$. \\
2) A verifier $V$, that takes the activations $\allx$ produced by the predictor and the corresponding label $y$ as input and produces as output the dual variables $\lambdall$. The verifier is parameterized by $\wparam$.

The training objective for PVT (Algorithm \ref{alg:PVT}) can be stated as follows:
\begin{align}
\mini_{\wall, \wparam} \quad & E_{(x, y) \sim \mathbb{P}}\left[(1 - 
\kappa) \ell\br{ P(x; \wall), y} + \kappa\log\br{1 + \dualfun\br{V\br{\allx, y ; \wparam};x, \wall}}\right].\label{eq:loss}
\end{align}
where $\ell$ is the supervised learning loss function (\eg, cross-entropy) and $\dualfun$ is the dual objective function from \eqref{eq:dualfun}, $\kappa$ is a hyperparameter that governs the relative weight of satisfying the specification versus fitting the data and $\mathbb{P}$ is the data generating distribution. Based on the empirical risk minimization principle \cite{vapnik2013nature}, this can be approximated with an empirical average over training samples.

This stands in contrast to prior work, where the dual variables for each training example are obtained by solving an optimization problem. Solving an optimization problem for every training example is a significant overhead even if it is only done on a minibatch of examples in every step of training. In \citet{kolter2017provable}, this computational overhead is alleviated to some degree by posing the gradient descent required to solve the dual optimization problem as backpropagation. However, despite this convenient reformulation, one still needs to solve an optimization problem for every example in the minibatch. The approach in \citet{CertifiedDefenses} sidesteps this by computing bounds that are valid for \emph{any training example}, but this approach is limited to single hidden layer networks and still requires solving a semidefinite program for each pair of classes, which is still computationally intensive.

PVT enables to overcome these limitations and scale up verified training to much larger models than that of previous approaches, while also making the training much faster even when compared to standard (non-verified) adversarial training \citep{madry2018towards}. Furthermore, as the verification bound is computed simply via a forward pass through the verifier network, the computational complexity of the verification procedure is linear in the size of the network. Since modern image classification architectures can contain millions of neurons, the asymptotic improvement obtained through learning is necessary to allow scalable verification and training of verifiable models.

\begin{figure*}[t]
\centering
\subfloat[Direct]{
    \resizebox{0.44\textwidth}{!}{
        \input{images/architectures/forward}
        \label{fig:forward_model}}}
\subfloat[Backward-forward]{
    \resizebox{0.54\textwidth}{!}{
        \input{images/architectures/backward_forward}
        \label{fig:backward_forward_model}}}
\caption{Examples of predictor-verifier network architectures.}
\label{fig:arch}
\end{figure*}
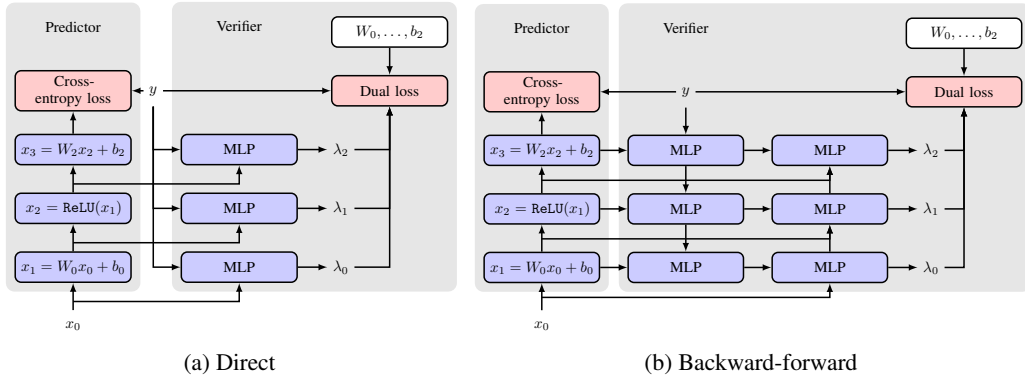

\subsection{Architecture of the verifier}
\label{sec:architecture}

We consider various choices for the architecture of the verifier network:

\paragraph{Constant verifier:} As the simplest option, we consider a trivial architecture that always predicts zero dual variables: $V\br{\allx,y;\theta}=0$ for any $\allx, y$. In this case, the verifier has no trainable parameters. This is equivalent to naive estimation of activation bounds based on simple bound propagation (Section \ref{sec:dual_verification}), which lacks the ability to model dependencies across activations.

\paragraph{Direct architecture:} The direct architecture predicts the dual variables independently for each layer, \ie, for layer $k$ it digests its inputs $x_k$ and the final target label $y$, and predicts the corresponding dual variable $\lambda_k$. This architecture predicts independently dual variables for each layer-wise constraint of the original problem~\eqref{eq:verifyOpt}. Figure~\ref{fig:forward_model} shows an example of such an architecture on a predictor network composed of two hidden layers and a final linear layer that outputs logits. We can see that each set of duals $\lambda_k$ are predicted from the corresponding layer inputs and the target label.

\paragraph{Backward-forward architecture:} The dual optimization \eqref{eq:verifyOptdual} can be viewed as a dynamic programming problem, where the stages of the dynamic program are layers in the neural network. The details are presented in Appendix \ref{sec:app_fb}. This can be used to derive a \emph{backward-forward architecture} for the verifier, illustrated in Figure~\ref{fig:backward_forward_model}. This architecture uses the output $x_{k+1}$ of each layer $k$ and propagates information downwards to create an intermediate representation (corresponding to the backward pass in a dynamic program). Each component of the backward pass extracts relevant information that is used in a subsequent forward pass. Finally, the forward pass takes the input of the each layer $x_k$ and the intermediate information computed by the backward pass to predict the dual variables $\lambdall$.

%% file: images/architectures/forward.tex
\begin{tikzpicture}

\node (input) [inp] {$x_0$};

\path (input.north)+(0, 1) node (layer_0) [mlp] {$x_1 = W_0 x_0 + b_0$};
\path (layer_0.north)+(0, 1) node (layer_1) [mlp] {$x_2 = \mathtt{ReLU}(x_1)$};
\path (layer_1.north)+(0, 1) node (layer_2) [mlp] {$x_3 = W_2 x_2 + b_2$};
\path (layer_2.north)+(0, 1) node (loss_0) [loss] {Cross-entropy loss};
\path (loss_0.east)+(.5, 0) node (label) [inp] {$y$};
\path [draw, ->] (input.north) -- node [above] {} (layer_0.south);
\path [draw, ->] (layer_0.north) -- node [above] {} (layer_1.south);
\path [draw, ->] (layer_1.north) -- node [above] {} (layer_2.south);
\path [draw, ->] (layer_2.north) -- node [above] {} (loss_0.south);
\path [draw, ->] (label.west) -- node [above] {} (loss_0.east);
\path (loss_0.north)+(0, 1) node (predictor_label) [inp] {Predictor};

\path (layer_0.east)+(2.5, 0) node (verifier_00) [mlp] {MLP};
\path (layer_1.east)+(2.5, 0) node (verifier_10) [mlp] {MLP};
\path (layer_2.east)+(2.5, 0) node (verifier_20) [mlp] {MLP};
\path (verifier_00.east)+(1, 0) node (dual_0) [inp] {$\lambda_0$};
\path (verifier_10.east)+(1, 0) node (dual_1) [inp] {$\lambda_1$};
\path (verifier_20.east)+(1, 0) node (dual_2) [inp] {$\lambda_2$};
\path (verifier_20.north)+(0, 2.5) node (verifier_label) [inp] {Verifier};
\path [draw, ->] (input.north)+(0, .2) -| node [] {} (verifier_00.south);
\path [draw, ->] (layer_0.north)+(0, .2) -| node [] {} (verifier_10.south);
\path [draw, ->] (layer_1.north)+(0, .2) -| node [] {} (verifier_20.south);
\path [draw, ->] (verifier_00.east) -- node [] {} (dual_0.west);
\path [draw, ->] (verifier_10.east) -- node [] {} (dual_1.west);
\path [draw, ->] (verifier_20.east) -- node [] {} (dual_2.west);
\path (loss_0.east)+(6, 0) node (loss_1) [loss] {Dual loss};
\path [draw, ->] (dual_0.east) -| node [above] {} (loss_1.south);
\path [draw, ->] (dual_1.east) -| node [above] {} (loss_1.south);
\path [draw, ->] (dual_2.east) -| node [above] {} (loss_1.south);
\path [draw, ->] (label.east) -- node [above] {} (loss_1.west);

\path (loss_1.north)+(0, 1) node (vars) [var] {$W_0, \ldots, b_2$};
\path [draw, ->] (vars.south) -| node [above] {} (loss_1.north);

\path [draw, ->] (label.south) |- node [above] {} (verifier_00.west);
\path [draw, ->] (label.south) |- node [above] {} (verifier_10.west);
\path [draw, ->] (label.south) |- node [above] {} (verifier_20.west);

\begin{pgfonlayer}{background}
\filldraw [line width=4mm,join=round,black!10]
  (predictor_label.north  -| loss_0.east) rectangle (layer_0.south -| layer_0.west)
  (verifier_label.north  -| verifier_00.west) rectangle (layer_0.south -| loss_1.east);
\end{pgfonlayer}

\end{tikzpicture}

%% file: images/architectures/backward_forward.tex
\begin{tikzpicture}

\node (input) [inp] {$x_0$};

\path (input.north)+(0, 1) node (layer_0) [mlp] {$x_1 = W_0 x_0 + b_0$};
\path (layer_0.north)+(0, 1) node (layer_1) [mlp] {$x_2 = \mathtt{ReLU}(x_1)$};
\path (layer_1.north)+(0, 1) node (layer_2) [mlp] {$x_3 = W_2 x_2 + b_2$};
\path (layer_2.north)+(0, 1) node (loss_0) [loss] {Cross-entropy loss};
\path (loss_0.east)+(2, 0) node (label) [inp] {$y$};
\path [draw, ->] (input.north) -- node [above] {} (layer_0.south);
\path [draw, ->] (layer_0.north) -- node [above] {} (layer_1.south);
\path [draw, ->] (layer_1.north) -- node [above] {} (layer_2.south);
\path [draw, ->] (layer_2.north) -- node [above] {} (loss_0.south);
\path [draw, ->] (label.west) -- node [above] {} (loss_0.east);
\path (loss_0.north)+(0, 1) node (predictor_label) [inp] {Predictor};

\path (layer_0.east)+(2, 0) node (verifier_00) [mlp] {MLP};
\path (layer_1.east)+(2, 0) node (verifier_10) [mlp] {MLP};
\path (layer_2.east)+(2, 0) node (verifier_20) [mlp] {MLP};

\path (verifier_00.east)+(2, 0) node (verifier_01) [mlp] {MLP};
\path (verifier_10.east)+(2, 0) node (verifier_11) [mlp] {MLP};
\path (verifier_20.east)+(2, 0) node (verifier_21) [mlp] {MLP};

\path [draw, ->] (verifier_20.south) -- node [above] {} (verifier_10.north);
\path [draw, ->] (verifier_10.south) -- node [above] {} (verifier_00.north);

\path (verifier_01.east)+(1, 0) node (dual_0) [inp] {$\lambda_0$};
\path (verifier_11.east)+(1, 0) node (dual_1) [inp] {$\lambda_1$};
\path (verifier_21.east)+(1, 0) node (dual_2) [inp] {$\lambda_2$};

\path (verifier_20.north)+(0, 2.5) node (verifier_label) [inp] {Verifier};

\path [draw, ->] (layer_0.east) -- node [] {} (verifier_00.west);
\path [draw, ->] (layer_1.east) -- node [] {} (verifier_10.west);
\path [draw, ->] (layer_2.east) -- node [] {} (verifier_20.west);

\path [draw, ->] (verifier_00.east) -- node [] {} (verifier_01.west);
\path [draw, ->] (verifier_10.east) -- node [] {} (verifier_11.west);
\path [draw, ->] (verifier_20.east) -- node [] {} (verifier_21.west);

\path [draw, ->] (verifier_01.north) -- node [] {} (verifier_11.south);
\path [draw, ->] (verifier_11.north) -- node [] {} (verifier_21.south);

\path [draw, ->] (input.north)+(0, .3) -| node [] {} (verifier_01.south);
\path [draw, ->] (layer_0.north)+(0, .3) -| node [] {} (verifier_11.south);
\path [draw, ->] (layer_1.north)+(0, .3) -| node [] {} (verifier_21.south);

\path [draw, ->] (verifier_01.east) -- node [] {} (dual_0.west);
\path [draw, ->] (verifier_11.east) -- node [] {} (dual_1.west);
\path [draw, ->] (verifier_21.east) -- node [] {} (dual_2.west);
\path (loss_0.east)+(8.5, 0) node (loss_1) [loss] {Dual loss};
\path [draw, ->] (dual_0.east) -| node [above] {} (loss_1.south);
\path [draw, ->] (dual_1.east) -| node [above] {} (loss_1.south);
\path [draw, ->] (dual_2.east) -| node [above] {} (loss_1.south);
\path [draw, ->] (label.east) -- node [above] {} (loss_1.west);

\path (loss_1.north)+(0, 1) node (vars) [var] {$W_0, \ldots, b_2$};
\path [draw, ->] (vars.south) -- node [above] {} (loss_1.north);

\path [draw, ->] (label.south) -- node [above] {} (verifier_20.north);

\begin{pgfonlayer}{background}
\filldraw [line width=4mm,join=round,black!10]
  (predictor_label.north  -| loss_0.east) rectangle (layer_0.south -| layer_0.west)
  (verifier_label.north  -| verifier_00.west) rectangle (layer_0.south -| loss_1.east);
\end{pgfonlayer}

\end{tikzpicture}

%% file: experiments.tex
\section{Experimental results}

We demonstrate that PVT can train verifiable networks, compare to the state-of-the-art, and discuss the impact of the verifier architecture on the results. For all datasets, we use a ConvNet architecture, comparable to the one used in \citet{kolter2017provable}: the input image is passed through two convolutional layers (with 16 and 32 channels and strides of 1 and 2, respectively), followed by two fully connected layers stepping down to 100 and then 10 (the output dimension) hidden units; ReLUs follow each layer except the last. Further details on the implementation and parameters are available in Appendix \ref{sec:app_implementation}.

\subsection{PVT and state-of-the-art}

Experimentally, we observe that either the direct or the backward-forward architecture achieve the best verified bounds on adversarial error rates. We compare these best-performing models (\ie~using tuned $\kappa$ values) to three alternative approaches. The baseline is the network trained in a standard manner with just the cross-entropy loss. The second algorithm is adversarial training, following \citet{madry2018towards}, which generates adversarial examples on the fly during training and adds them to the training set. The third algorithm is from \citet{kolter2017provable} -- we only perform this comparison on MNIST and SVHN benchmarks, as they do not test their approach on CIFAR-10.

Table~\ref{table:results} contains the results of these comparisons. Compared to \citet{kolter2017provable}, our predictor-verifier training (PVT) achieves lower error rates under normal and adversarial conditions, as well as better verifiable bounds, setting the state-of-the-art in verified robustness to adversarial attacks.
PVT outperforms adversarial training \citep{madry2018towards}, achieving lower error rates against adversarial attacks of 2.87\% on MNIST and 67.28\% on CIFAR-10 for this particular ConvNet architecture.
In addition, our networks are provably robust while the adversarially trained networks only achieve the trivial upper bound of 100\% adversarial error rate - \emph{this shows that training networks to be verifiable is critical to obtaining networks with good verified bounds on adversarial accuracy}. However, our networks do have much higher nominal test error (without adversarial perturbations), making space for improvement in future work. We note that on CIFAR-10 our trained network is the first to achieve a non-trivial verified upper bound on adversarial error. Finally, PVT takes 6 minutes to reach the same performance as \cite{kolter2017provable} on MNIST. \citet{kolter2017provable} take 5 hours to train their model on MNIST, using a slightly weaker GPU (Titan X versus our Titan Xp). To reach our best performance, we take 20 minutes. Compensating for the difference in hardware, we are roughly 25$\times$ faster.
\begin{table}[tb]
\caption{{\bf Comparison with the state-of-the-art.}
Methods are evaluated in terms of the test error (no attack),
the currently strongest PGD attack \cite{madry2018towards},
and the provable upper bound on the adversarial error. $\epsilon$ is the size of the $\ell_\infty$ attack for pixel values in $[0, 1]$, \eg, $\epsilon=0.03$ corresponds roughly to an 8 pixel attack when pixel values are in $[0, 255]$. All methods use comparable ConvNet architectures. *For comparison, the last line shows the best achievable adversarial accuracy using a much larger model on CIFAR-10.
}
\vspace{.3cm}
\label{table:results}
\centering
\begin{tabular}{llcrrr}
\hline
{\sc Problem} & {\sc Method} & $\epsilon$ & {\sc Test Error} & {\sc PGD Attack} & {\sc Bound} \\
\hline
{\sc MNIST} & Baseline & 0.1 & 0.77\% & 52.94\% & 100.00\% \\
{\sc MNIST} & \citet{kolter2017provable} & 0.1 & 1.80\% & 4.11\% & 5.82\% \\
{\sc MNIST} & \citet{madry2018towards} & 0.1 & {\bf 0.60\%} & 4.66\% & 100.00\% \\
{\sc MNIST} & Predictor-Verifier & 0.1 & 1.20\% & {\bf 2.87\%} & {\bf 4.44\%} \\
\hline
{\sc SVHN} & Baseline & 0.01 & {\bf 6.57\%} & 87.45\% & 100.00\% \\
{\sc SVHN} & \citet{kolter2017provable} & 0.01 & 20.38\% & 33.74\% & 40.67\% \\
{\sc SVHN} & \citet{madry2018towards} & 0.01 & 7.04\% & {\bf 23.63\%} & 100.00\% \\
{\sc SVHN} & Predictor-Verifier & 0.01 & 16.59\% & 33.14\% & {\bf 37.56\%} \\
\hline
{\sc CIFAR-10} & Baseline & 0.03 & {\bf 26.27\%} & 99.99\% & 100.00\% \\
{\sc CIFAR-10} & \citet{madry2018towards} & 0.03 & 39.00\% & 68.08\% & 100.00\% \\
{\sc CIFAR-10} & Predictor-Verifier & 0.03 & 51.36\% & {\bf 67.28\%} & {\bf 73.33\%} \\
{\sc CIFAR-10} & \citet{madry2018towards}* & 0.03 & 12.7~\% & 54.2~\% & 100.00\% \\
\hline
\end{tabular}
\end{table}

\vspace{-.2cm}
\subsection{Verification time and importance of verified training}
\begin{figure*}[t]
\centering
\subfloat[PVT model: Verified error rate versus perturbation radius \label{fig:Verified_bounds_a}]{
    \includegraphics[width=.45\textwidth]{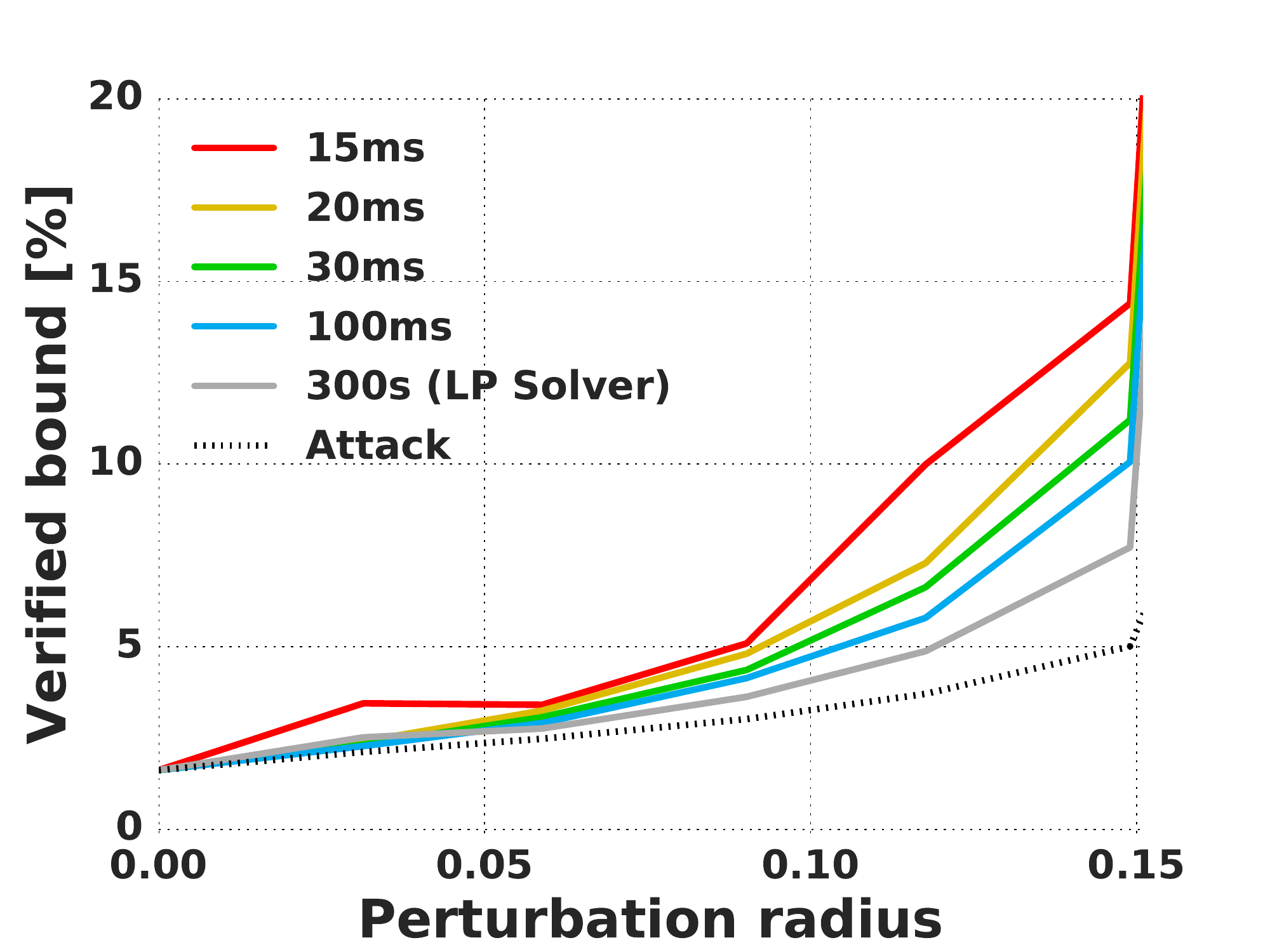}}
    \hspace{1cm}
\subfloat[Verification time: PVT model versus Adversarially trained model \label{fig:Verified_bounds_b}]{
    \includegraphics[width=.45\textwidth]{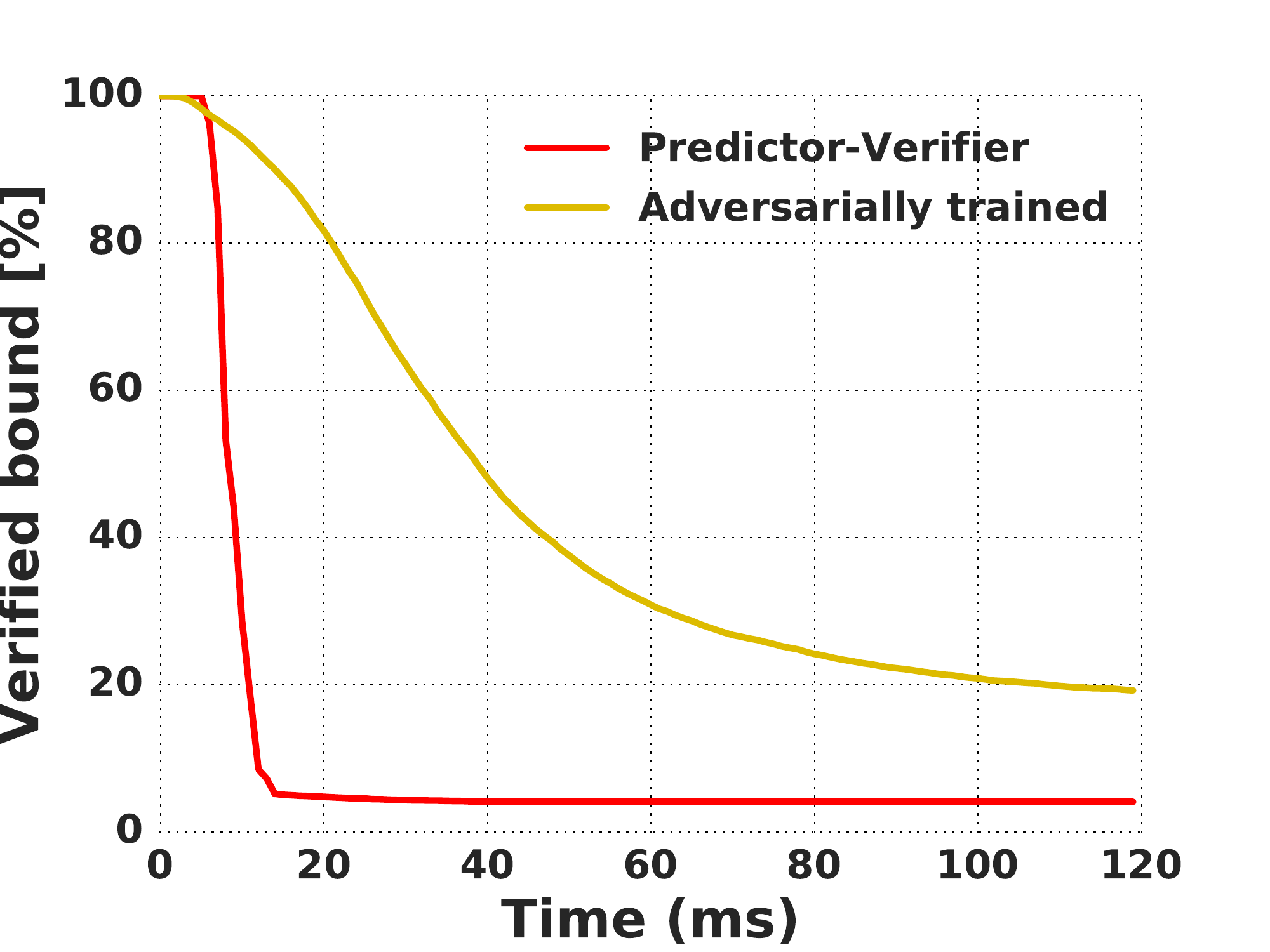}}
\caption{\protect\subref{fig:Verified_bounds_a} The effect of the perturbation radius on the verified bound for the MNIST PVT model. \protect\subref{fig:Verified_bounds_b} The decrease in the verified upper bound on MNIST adversarial error rate as a function of time taken by the verification algorithm (per test example) for the PVT and \citet{madry2018towards} models.}
\label{fig:Verified_bounds}
\end{figure*}

We also compare our models with those produced by the approach from \citet{madry2018towards} in terms of \emph{verifiability}. We were unable to obtain convolutional models using adversarial training that could be verified to obtain non-trivial bounds on adversarial accuracy, hence we resorted to using smaller fully connected models trained with L1 regularization, for which we were able to obtain non-trivial verified bounds on adversarial accuracy. The results shown in Figure \ref{fig:Verified_bounds} illustrate two important points: (a) The models trained with PVT are \emph{very easy to verify}, where with a budget of 15ms per test example, we are able to achieve nearly optimal bounds on the adversarial error rate, which can be seen from the gap between the solid lines (verified bounds) and dashed grey lines (CW attack) (Figure \ref{fig:Verified_bounds_a}); and
(b) The models trained with adversarial training, even with tuned $L_1$ regularization, achieve weaker verified bounds than the PVT model for any limit on verification time (Figure \ref{fig:Verified_bounds_b}).
\subsection{Comparison of verifier architectures in PVT}
\label{sec:res_architecture}

Table~\ref{table:arch} compares the performance of different verifier architectures. For this table and to facilitate comparisons, we set $\kappa = 1$ for all models (hence we expect the nominal error rate to increase w.r.t.\ Table~\ref{table:results}). The last column of the table corresponds to the verified bound obtained directly using the dual variables predicted by the verifier network, hence it is higher than the fully verified bounds shown in Table~\ref{table:results}. As expected, the constant architecture performs the worst as it systematically gets higher nominal and verified error rates. The direct architecture is competitive against the backward-forward architecture. However, on SVHN the difference is more pronounced as the backward-forward architecture achieves a much lower nominal error rate.

\begin{table}[t]
\caption{{\bf Comparison of verifier architectures.} Methods are evaluated in terms of the test error (no attack), the currently strongest PGD attack \cite{madry2018towards}, and the verified bound on adversarial error, obtained directly using dual variables predicted by the verifier network. For this table, all models use $\kappa = 1$.}
\vspace{.3cm}
\label{table:arch}
\centering
\begin{tabular}{llcrrr} 
\hline
{\sc Problem} & {\sc Architecture} & $\epsilon$ & {\sc Test Error} & {\sc PGD Attack} & {\sc Bound}* \\
\hline
{\sc MNIST} & Constant & 0.1 & 1.29\% & 3.08\% & 5.00\% \\
{\sc MNIST} & Direct & 0.1 & {\bf 1.20\%} & {\bf 2.87\%} & {\bf 4.44\%} \\
{\sc MNIST} & Backward-forward & 0.1 & 1.28\% & 3.07\% & 4.72\% \\
\hline
{\sc SVHN} & Constant & 0.01 & 24.80\% & 38.03\% & 45.92\% \\
{\sc SVHN} & Direct & 0.01 & 22.50\% & 35.98\% & 42.93\% \\
{\sc SVHN} & Backward-forward & 0.01 & {\bf 16.59\%} & {\bf 33.14\%} & {\bf 41.52\%} \\
\hline
{\sc CIFAR-10} & Constant & 0.03 & 60.12\% &69.32\% & 72.21\% \\
{\sc CIFAR-10} & Direct & 0.03 & 59.38\% & {\bf 67.68\%} & {\bf 70.79\%} \\
{\sc CIFAR-10} & Backward-forward & 0.03 & {\bf 58.89\%} & 68.05\% & 71.36\% \\
\hline
\end{tabular}
\end{table}

%% file: conclusions.tex
\vspace{-3mm}
\section{Conclusions}
\vspace{-1mm}
We have presented a novel approach for training of verified models -- we believe that this is a significant step towards the vision of specification-driven ML. Our experiments have shown that the proposed predictor-verified approach  outperform competing techniques in terms of tightness of the verified bounds on adversarial error rates and nominal error rates in image classification problems, while also training faster. In the future, we plan to apply the predictor-verifier framework to bigger datasets like ImageNet and to extend it to handle complex specifications encountered in real-world machine learning tasks.

%% file: appendix.tex
\newpage
\appendix
\section{Appendix}
\subsection{Closed form dual objective function} \label{sec:AppCF}
This derivation mimics that in \citep{DjDual} closely, but we reproduce it here for completeness. Our goal is to show that the optimization problem
\begin{align}
\begin{split}
\maxi \quad & c^Tx_K + d + \sum_{k=0}^{K-1} \lambda_k^T(x_{k+1} - h_k(x_k))\\
\mbox{subject to} \quad & x_0 \in \Sin\br{x^\text{nom}} \\
&
l_k \leq x_k \leq u_k, \quad k = 1, \ldots, K
\end{split}
\end{align}
over variable $\allx$
can be solved efficiently for fixed $\lambdall$, where we assume for ease of exposition that $\Sin(\xnom) = \{x \mid \|x - \xnom\|_\infty \leq \epsilon\}$ for some $\epsilon \geq 0$. We begin by rearranging the objective:
\begin{align}
\begin{split}
\maxi \quad & {\br{c+\lambda_{K-1}}}^Tx_K + d + \sum_{k=1}^{K-1} \br{{\br{\lambda_{k-1}}}^Tx_k - \lambda_k^T h_k(x_k)}-\lambda_0^Th_0(x_0)\\
\mbox{subject to} \quad & \xnom - \epsilon \leq x_0 \leq \xnom+\epsilon \\
&
l_k \leq x_k \leq u_k, \quad k = 1, \ldots, K.
\end{split}
\end{align}
The objective function is \emph{separable}, \ie, is a sum of terms each of which depends on a single $x_k$. The constraints are also separable, \ie, each constraint only involves a single $x_k$. Thus, the optimization can be solved independently for each $x_k$. For $k=1, \ldots, K-1$, we have
\[\maxi_{l_k \leq x_k \leq u_k} {\br{\lambda_{k-1}}}^Tx_k - \lambda_k^T h_k(x_k),\]
and due to our assumption that the transfer functions $h_k$ are simple, this optimization problem can be solved efficiently. We denote the optimal value $f_k\br{\lambda_{k-1}, \lambda_k}$.
For $k=0$, we have
\[
\maxi_{\|x_0-\xnom\|_\infty\leq \epsilon} \quad {-\lambda_{0}}^Th_0(x_0)\]
which can be solved similarly; we denote the optimal value $f_0\br{\lambda_0}$.
For $k=K$, we have
\[
\maxi_{l_K \leq x_K \leq u_K} \quad {\br{c+\lambda_{K-1}}}^Tx_K\]
which can be solved in closed form by setting each component of $x_K$ to be its upper bound ($u_K$) or lower bound ($l_K$) depending on whether the corresponding component of $c+\lambda_K$ is positive or not. We denote the optimal solution by $f_K\br{\lambda_{K-1}}$. Thus, the dual objective can be written as
\begin{align}
f_0\br{\lambda_0} + f_{K}\br{\lambda_{K-1}} +  \sum_{k=1}^{K-1} f_k\br{\lambda_{k-1}, \lambda_k}.
\label{eq:DualFin}
\end{align}

\subsection{Derivation of backward-forward architecture}\label{sec:app_fb}

This architecture of the verifier is motivated by the structure of the dual function \eqref{eq:DualFin}. Each term in the objective function depends on a single $\lambda_k$ or on a pair of adjacent $\lambda_{k-1}, \lambda_k$. The dual optimization problem is to minimize eq.~(\ref{eq:DualFin}) over $\lambdall$. This can be solved as a dynamic program. The recursion proceeds as follows: compute the value function $F_{K-1}(\lambda_{K-1}) = f_{K-1}\br{\lambda_{K-1}}$, then for $k=K-2, \ldots, 0$ compute
\begin{equation}
F_k(\lambda_k) = \min_{\lambda_{k+1}} f_{k+1}\br{\lambda_k, \lambda_{k+1}} + F_{k+1}\br{\lambda_{k+1}}.
\label{eq:DPback}
\end{equation}
The optimal value is then given by $\min_{\lambda_0} F_0\br{\lambda_0}$ and the optimal solutions can be computed by starting from $\lambda_0^\star =\argmin_{\lambda_0} F_0\br{\lambda_0}$ and then for $k=1, \ldots, K-1$ calculating
\begin{equation}
\label{eq:DPforward}
\lambda_{k}^\star =\argmin_{\lambda_{k}} f_{k}\br{\lambda_{k-1}^\star, \lambda_{k}} + F_{k}\br{\lambda_{k}}.
\end{equation}
While this procedure can be described in theory, it may not necessarily be implementable. This is because the value functions \eqref{eq:DPback} may not have a closed form expression. Thus, it may not possible to represent them using a finite amount of computation or storage. A standard approach to overcome this problem is to use a parametric approximation of the value functions $F_k$.
However, our goal in this paper is not to approximate the value functions, but use to it to construct an architecture that can produce $\lambdall$ given activations $\allx$ and label $y$. In doing this, we try to mimic the computational graph structure.
Let $G_k$ and $E_k$, $k=0, \ldots, K-1$, be (typically small) neural networks that we will use to approximate the backward-forward computations of (\ref{eq:DPback}) and (\ref{eq:DPforward}). Then starting with $\eta_{K-1} = G_{K-1}\br{x_{K}, y}$, we compute activations $k=K-2, \ldots, 0$ as
\begin{equation}
\eta_{k} = G_k\br{\eta_{k+1}, x_{k+1}}.
\label{eq:DPbackfin}
\end{equation}
This computation runs backward through the layers of the network to produce $\eta_0$. Then, the $\lambdall$ can be computed as follows, starting from $\lam{0}= E_0\br{\eta_0, x_0}$, we compute for $k=0, \ldots, K-1$
\begin{equation}
\label{eq:DPforwardfin}
\lambda_{k} = E_k(\lambda_{k-1}, \eta_k).
\end{equation}
The systems of equations \eqref{eq:DPbackfin} and \eqref{eq:DPforwardfin} together constitute a mapping from $(\allx , y)$ to $\lambda$ and can be trained end-to-end to minimize the dual bound. This structure is represented by the backward-forward architecture described in section \ref{sec:architecture}. We can denote this mapping
\[\lambdall = T\br{\allx, y; \theta}\]
where $\theta$ is the set of all parameters of the `dual' networks $G_k$ and $E_k$, $k=0, \ldots, K-1$. As long as $G$ and $E$ are differentiable, so is $T$ and its derivatives can be computed using backpropagation. Note that since the activations $\allx$ depends on the weights of the predictor network $\wall$ we can backpropagate the gradient of the loss both into the $\theta$ parameters and into the $\wall$ parameters.

\subsection{Implementation details}
\label{sec:app_implementation}

Here we provide additional implementation details on the model architectures, training procedure, and how we improved the speed of bound propagation.

\paragraph{Details of verifier architectures.}
For the direct architecture, we predict each set of dual variables using two fully connected layers stepping down to 200 and then up to the required number of dual variables (the output dimension); ReLU follows the first layer.
For the backward-forward architecture, the backward path for each set of dual variables is a single layer of size 200 followed by a ReLU, and the forward path uses two fully connected layers stepping down to 200 and then up to the required number of dual variables.

\paragraph{Datasets and training data augmentation.}
Throughout the paper, we concentrate on three widely used datasets: MNIST~\cite{mnist}, SVHN~\cite{netzer2011reading} and CIFAR-10~\cite{krizhevsky2009learning}.
For MNIST, following standard practice, no data augmentation is performed, apart from for the adversarial training baseline \citet{madry2018towards} which augments the training data with adversarial examples.
For SVHN, we train on random $28\times28$ crops of images in the training set (without \emph{extras}) and evaluate on $28\times28$ central crops of the test set.
For CIFAR-10, we train on random $28\times28$ crops and horizontal flips, and evaluate on $28\times28$ central crops.

\paragraph{Training procedure.}
The networks were trained using the Adam \cite{kingma2015adam} algorithm with an initial learning rate of $10^{-3}$.
We train for 90, 80 and 240 epochs with batch sizes of 100, 50 and 50 on MNIST, SVHN and CIFAR-10 respectively.
The value of $\kappa$ was tuned independently for each dataset, but as we will see in Appendix~\ref{sec:pareto} there is a trade-off between nominal and adversarial accuracy. Ultimately, for MNIST we chose $\kappa = 1$, for SVHN $\kappa = 1$ and for CIFAR-10 $\kappa = 0.5$.
We also found it necessary to linearly anneal the perturbation radius $\epsilon$ during the first half of the training steps between $0$ and the required value, \ie~$0.1$, $0.01$ and $0.03$ on MNIST, SVHN and CIFAR-10, respectively.
Finally, we added $L_1$-regularization to the predicted dual variables with scale $10^{-6}$ in order to force the verifier to concentrate on a smaller subset of dual variables.

\paragraph{Efficient bound propagation for affine layers.}
As explained in Section \ref{sec:dual_verification}, given bounds on the inputs $l \leq x \leq u$ to a transfer function $h$, it is useful to be able to efficiently compute bounds on the outputs $l' \leq h(x) \leq u'$. For exposition clarity, the main paper presents a suboptimal way of propagating the bounds for an affine function $h(x) = Wx + b$ as follows:
\[
\begin{array}{rcl}
l^\prime &=& W^+ l + W^- u + b \\
u^\prime &=& W^+ u + W^- l + b \\
\end{array}
\]
where $W^+=\max\br{W, 0}$ and $W^-=\min\br{W, 0}$.

In fact, it is possible to halve the number of operations needed to compute the output bounds -- let us reparametrize bounds by representing them as centre $c=(l+u)/2$ and radius $r=(u-l)/2$; conversely $l=c-r$ and $u=c+r$. Then, the output bounds are simply

\[
\begin{array}{rcl}
c^\prime &=& W c + b \\
r^\prime &=& \lvert W \rvert r \\
\end{array}
\]

Therefore, only two affine operations are required to propagate bounds for affine transfer functions, compared to four of the original parametrization.
Switching between parametrizations depending on $h$ (for element-wise non-linearities $(l,u)$ is better, while for affine $(c,r)$ is more efficient) incurs a slight computational overhead, but since affine layers are typically the most computationally intensive ones (fully connected layers, convolutions, \etc), the $2\times$ speedup is worth it.

\subsection{Other experimental results}
\label{sec:pareto}
\begin{figure}[t]
\centering
\subfloat[MNIST ($\epsilon = 0.1$)]{
\includegraphics[width=.4\textwidth]{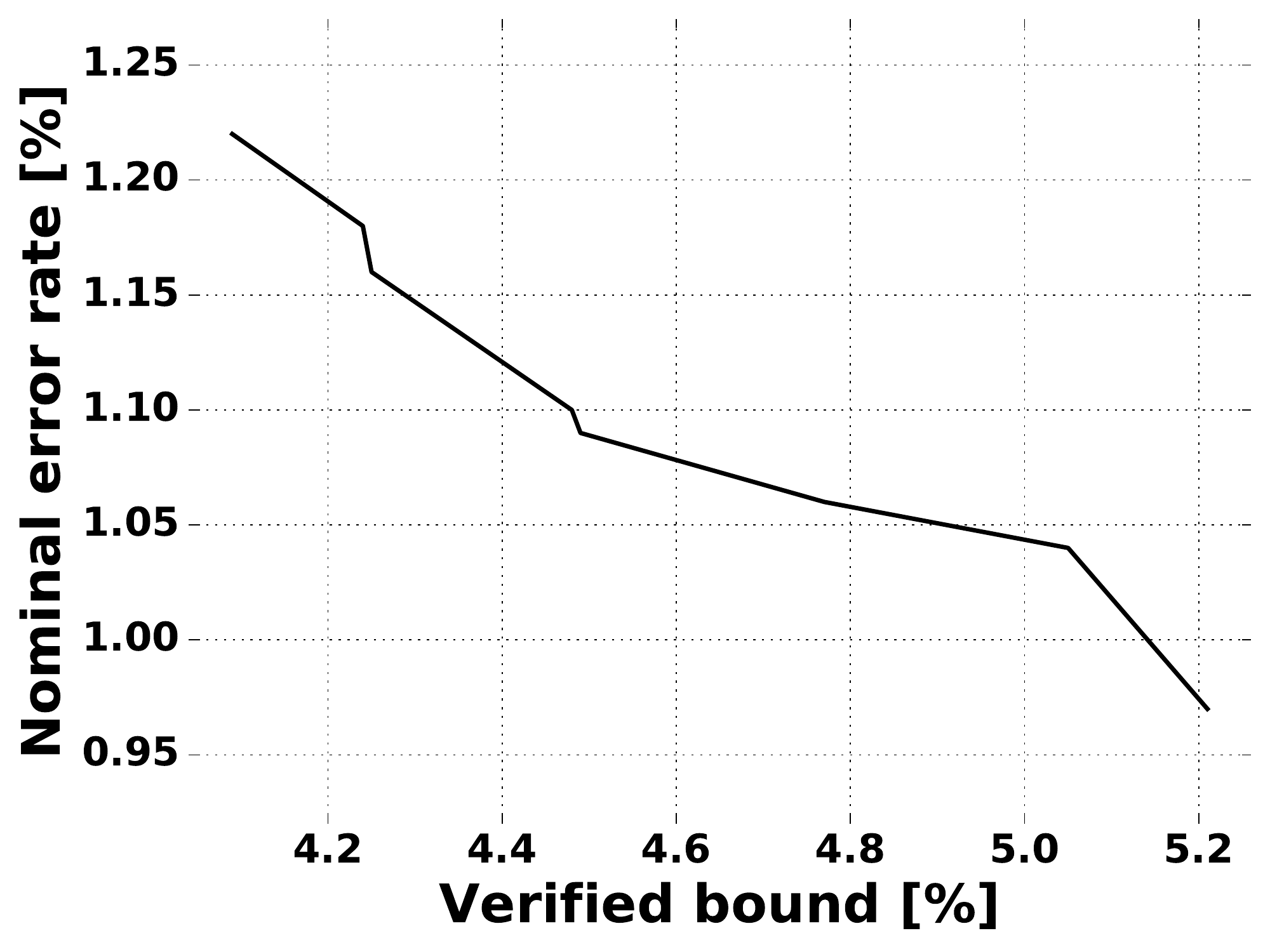}
\label{fig:pareto_mnist}
}
\hspace{1cm}
\subfloat[CIFAR-10 ($\epsilon = 0.03$)]{
\includegraphics[width=.4\textwidth]{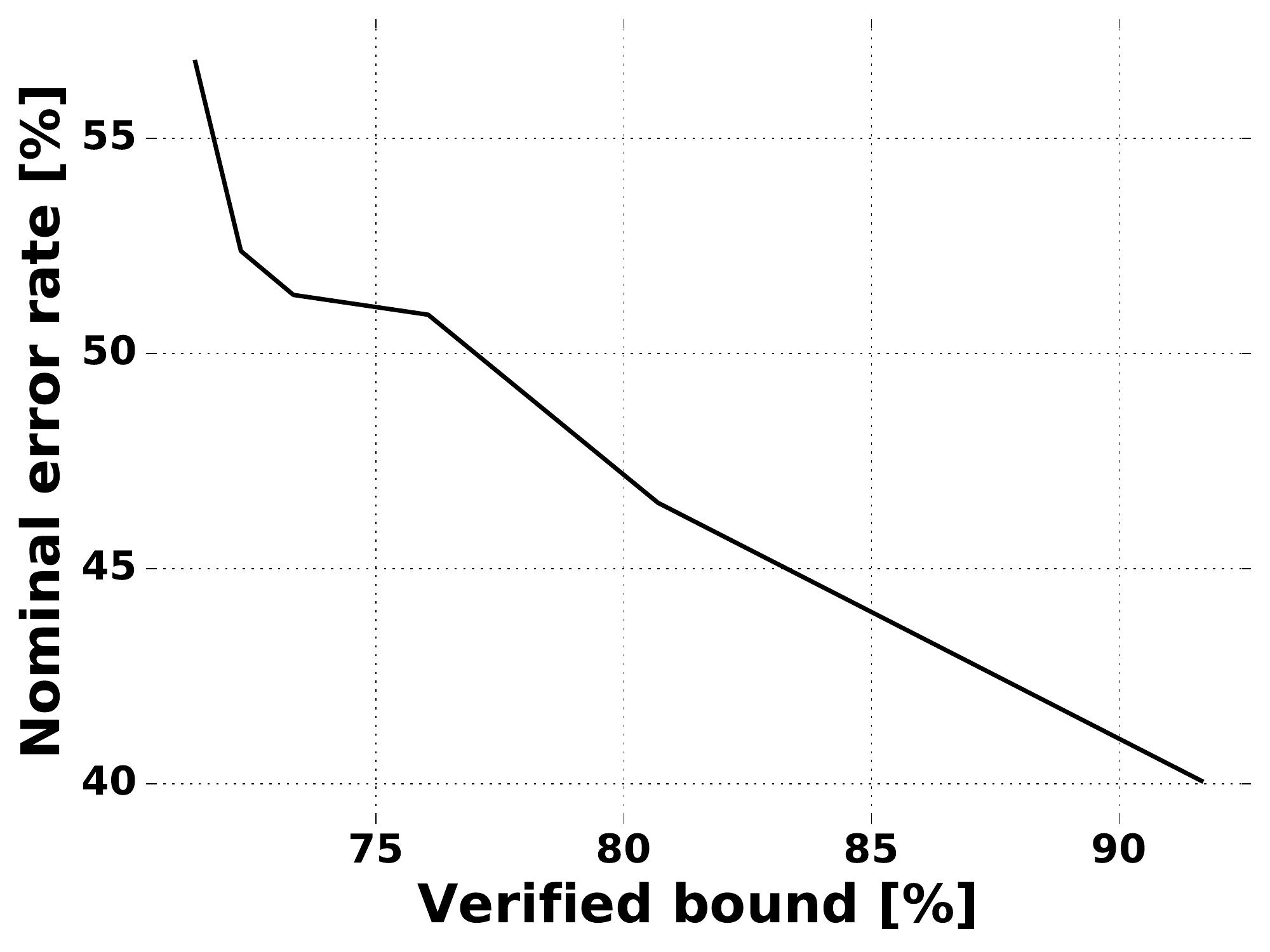}
\label{fig:pareto_cifar}
}
\caption{Trade-off between the nominal error rate (without any adversarial perturbations and the verified upper bound on \protect\subref{fig:pareto_mnist} MNIST with $\epsilon = 0.1$ and \protect\subref{fig:pareto_cifar} CIFAR-10 with $\epsilon = 0.03$}
\label{fig:pareto}
\end{figure}

Figure~\ref{fig:pareto} shows the effect of $\kappa$ on the nominal (without any adversarial perturbations) and verified error rate. For this plot, we trained multiple models with different values of $\kappa$ ranging between 0.05 and 1 on MNIST and CIFAR-10 and report the performance of these models at the end of training.
We observe a clear anti-correlation between the nominal error rate and the verified upper bound. Indeed, as we force the model to prefer the maximization of the nominal accuracy (\ie~by setting $\kappa$ close to zero), robustness takes a hit.

